\documentclass[twoside]{article}

\usepackage[accepted]{aistats2023}
\usepackage[round]{natbib}

\usepackage[utf8]{inputenc}

\usepackage[utf8]{inputenc} 
\usepackage[T1]{fontenc}    
\usepackage{url}            
\usepackage{booktabs}       
\usepackage{scalefnt}
\usepackage{nicefrac}       
\usepackage{microtype}      
\usepackage{optidef}
\usepackage{amsmath,amsfonts,amssymb}
\usepackage{bbm}
\usepackage{mathtools}
\usepackage{stmaryrd}
\usepackage{subcaption}
\usepackage{graphicx} 
\usepackage{bbm}
\usepackage[linesnumbered,ruled]{algorithm2e}
\SetAlCapNameFnt{\scriptsize}
\SetAlCapFnt{\scriptsize}
\usepackage[table,xcdraw]{xcolor}

\SetKwInput{KwInput}{Input}                
\SetKwInput{KwOutput}{Output}  
\SetKwRepeat{Repeat}{repeat}{until}
\SetKwRepeat{RepeatFor}{repeat}{for}
\SetKwRepeat{Repeat}{repeat}{until}

\SetCommentSty{mycommfont}

\begin{document}

\twocolumn[

\aistatstitle{\textsc{ForestPrune}: Compact Depth-Pruned Tree Ensembles}

\aistatsauthor{ Brian Liu \And Rahul Mazumder}

\aistatsaddress{ Operations Research Center \\
MIT \And   Sloan School of Management \& Operations Research Center\\
MIT} ]

\begin{abstract}

Tree ensembles are powerful models that achieve excellent predictive performances, but can grow to unwieldy sizes. These ensembles are often post-processed (pruned) to reduce memory footprint and improve interpretability. We present \textsc{ForestPrune}, a novel optimization framework to post-process tree ensembles by pruning depth layers from individual trees. Since the number of nodes in a decision tree increases exponentially with tree depth, pruning deep trees drastically compactifies ensembles. We develop a specialized optimization algorithm to efficiently obtain high-quality solutions to problems under \textsc{ForestPrune}. Our algorithm typically reaches good solutions in seconds for medium-size datasets and ensembles, with 10000s of rows and 100s of trees, resulting in significant speedups over existing approaches.
Our experiments demonstrate that \textsc{ForestPrune} produces parsimonious models that outperform models extracted by existing post-processing algorithms.
\end{abstract}

\section{INTRODUCTION}
Tree ensembles are popular in machine learning for their predictive accuracy and interpretability. These ensembles combine weak decision trees by either averaging independently grown trees in bagging \citep{breiman2001random}, or by adding trees grown sequentially in boosting \citep{friedman2001greedy}. The resulting model is more accurate and generalizes better than any single decision tree, but is much more complex. 

The complexity of tree ensembles raises several issues. Tree ensembles can grow to massive sizes and require substantial memory to store. Computing the predictions of large ensembles is slow since each observation in a dataset must pass through each tree. Ensembles with deep trees are also difficult to interpret since the structures of deep trees are hard to visualize \citep{haddouchi2019survey}. Finally, complex tree ensembles can overfit, and tuning large ensembles is time-consuming. To remedy these issues, tree ensembles can be post-processed after training to more parsimonious forms. Post-processing a tree ensemble improves memory footprint, prediction speed, and interpretability without retraining the model \citep{sagi2018ensemble}.

We present \textsc{ForestPrune}, a novel optimization framework for tree ensemble post-processing. Compared to traditional methods that post-process tree ensembles by selecting subsets of decision trees, \textsc{ForestPrune} trims the depths of trees in an ensemble. By removing layers, decision trees can be reduced in depth or eliminated, and as a result, the framework can prune shallow and sparse subensembles that perform well.  We develop a specialized block coordinate descent algorithm \citep{nutini2017let} to obtain high-quality solutions to problems in \textsc{ForestPrune} by minimizing regularized loss. Our specialized solver allows computation for instances that appear to be beyond the capabilities of off-the-shelf optimization solvers. Finally, we show in our experiments the advantages of trimming depth layers compared to removing trees across a range of scenarios, in terms of both improving prediction performance and interpretability. Our implementation of \textsc{ForestPrune} is open-source and publicly available via GitHub \footnote{https://github.com/brianliu12437/ForestPruneAISTATS2023}.

\subsection{Background and Related Work}

We give a brief overview of tree-based methods for regression problems. Decision trees form the base learners used in tree ensembles. A regression tree of depth $d$ partitions a dataset into at most $2^d$ non-overlapping partitions (leaf nodes). Within each leaf node, the prediction is the mean response of the observations in that partition. Decision trees are not necessarily balanced, but the number of nodes in a tree generally increases exponentially with tree depth. Bagging averages the predictions of many weakly correlated decision trees to reduce variance \citep{breiman2001random}. The method generally does not overfit with respect to the number of features used, the number of trees grown, or the depth of each tree \citep{maclin1997empirical}. Typically each tree is grown to full depth where each leaf node contains a single observation. Boosting sums the predictions of a sequence of decision trees, where each tree in the sequence is fit on the residuals of the prior ensemble. At each boosting iteration, the predictions of the current tree are dampened (i.e, shrunk) by the learning rate $\gamma$. This parameter controls the effect that each tree has on the ensemble as well as the similarity between trees adjacent in the boosting sequence.

We also provide an overview of existing work on post-processing tree ensembles to improve compactness. \cite{friedman2003importance} introduces the importance sampled learning ensemble (ISLE) framework. This framework involves growing a tree ensemble using various sampling techniques and post-processing the ensemble in the following manner. Assign each tree $t$ in the ensemble coefficient $\beta_t$. Incorporate an $\ell_1$-penalty on the weights, $\sum_t |\beta_t|$, to encourage shrinkage and sparsity. The ensemble produced is compact and often performs comparably to the full model. ISLE regularizes the number of trees in the ensemble as a proxy for ensemble size, the framework is unable to trim individual trees in the ensemble. \cite{friedman2008predictive} extends this approach by proposing $\ell_1$-regularization over the node space. While this approach is interesting, the number of nodes in a tree increases exponentially with tree depth, so the total number of nodes in a full depth ensemble is enormous. As a result, it is often infeasible to minimize $\ell_1$-regularized loss over the node space. Pruning nodes also destroys the tree structures in the ensemble, which hurts model interpretability. Besides regularization, various selection heuristics can be used to decide which trees to remove from an ensemble \citep{lucchese2016post,martinez2008analysis}. These heuristics select a compact subset of trees with respect to some target such as performance, diversity, or interpretability.


\section{\textsc{ForestPrune} Framework}
We detail \textsc{ForestPrune} for regression ensembles and assume without loss of generality that all trees in the ensemble are grown to the same maximum depth. Our main goal is to introduce an optimization framework to prune depth layers from trees in a trained ensemble. Using so-called depth-difference matrices, we show that this framework can be expressed as a regularized least-squares criteria, where the combinatorial penalty controls how many layers to prune. 

\noindent{\bf Notation } Given dataset $X$ with $m$ rows and $p$ columns, $X \in \mathbb{R}^{m \times p}$, and response $y \in \mathbb{R}^m $, tree $T_i$ maps $T_i(X) : \mathbb{R}^{m \times p} \rightarrow \mathbb{R}^{m}$. Let the prediction of $T_i(X) = \hat{y_i}$. A tree ensemble is a collection of $n$ trees $T_i, \ i \in [n]$ grown via bagging or boosting, and the prediction of the ensemble is given by $\hat{y} = \sum_{i=1}^{n} \gamma T_i(X)$. In the context of bagging, $\gamma = \frac{1}{n}$ and in the context of boosting $\gamma$ is the learning rate. Let $d$ denote the depth of trees in the ensemble.

\subsection{Depth-Difference Matrix}
For tree $T_i$, depth-difference matrix $D_i \in \mathbb{R}^{m \times d}$ encodes the decision path of each observation in $X$, with respect to the depth layers that the observations traverse. To initialize \textsc{ForestPrune} compute $D_i$ for each tree in the ensemble. 

Algorithm \ref{depthdiffalgo} presents the procedure for constructing $D_i$. Consider a single decision tree $T_i$ of depth $d$. For each observation $x_j \in X$, find the sequence of nodes in the tree traversed by $x_j$, $N_1 \shortrightarrow  N_2 \shortrightarrow \ldots \shortrightarrow N_k$. For each node in this sequence, compute the mean of the observations found in the node, $\mu_1  \shortrightarrow \mu_2  \shortrightarrow \ldots
 \shortrightarrow \mu_k$. The last element of this sequence $\mu_k$ is the prediction for observation $x_j$, $T_i(x_j)$. Take the rolling difference of values in this sequence of means and store these differences in vector $v_j$ where,
\[v_j = [ \mu_1, \mu_2 - \mu_1, \mu_3 - \mu_2, \ldots, \mu_k - \mu_{k-1}].\]
Pad the tail end of $v_j$ with zeros so that $v_j \in \mathbb{R}^{d}$. The vector $v_j$ has the property that $\langle v_j , \mathbf{1} \rangle = T_i(x_j)$, i.e., the elements of $v_j$ sum to the prediction of tree $T_i$ for observation $x_j$.  Repeat this procedure for $j \in [m]$ and store the results $v_j$ as the rows of $D_i \in \mathbb{R} ^{m \times d}$.

\begin{algorithm}
\scriptsize 
\caption{Computing Depth-Difference Matrices}
\label{depthdiffalgo}
\DontPrintSemicolon
  
  \KwInput{ Tree $T_i$ of depth $d$, $X \in \mathbb{R}^{m \times p}, y \in \mathbb{R}^m$}
  $D_i \leftarrow \{ \}$
  
  \For{$x_j \in X$}{
  
  Find sequence of nodes in $T_i$ traversed by $x_j$, $N_1 \shortrightarrow  N_2 \shortrightarrow \ldots \shortrightarrow N_k$.
 
  Compute the mean of $y$ partitioned by each node, $\mu_1  \shortrightarrow \mu_2  \shortrightarrow \ldots \shortrightarrow \mu_k$.
  
  Difference the sequence of means and store in vector $v_j = [\mu_1, \mu_2 - \mu_1, \ldots, \mu_k - \mu_{k-1}]$.
  
  Pad the tail of $v_j$ with zeros, $v_j \in \mathbb{R}^{d}$.
  
  Append $v_j$ as a row to $D_i$.
  }
  \KwOutput{$D_i \in \mathbb{R}^{m \times d}$}
\end{algorithm}

Depth-difference matrices are efficient to compute since the sequence of nodes each observation traverses, along with the corresponding $\mu$ values of each node, are obtained as a byproduct of the tree training process. Differencing the sequence of $\mu$ values requires $d$ operations, and since $d \ll m$, the cost of computing depth-difference matrices is linear in the number of training observations, $O(m)$.

Below, we introduce notation and functions for the vectors of binary decision variables, $z_i \in \{0,1\}^d, \ i \in [n]$,  that will appear in our optimization formulation. These variables work in conjunction with depth-difference matrices to prune tree ensembles.

\noindent{\bf Pruning Depth Layers }
Given depth-difference matrix $D_i$, the predictions of the full tree for all observations, $T_i(X) = \hat{y_i}$, can be obtained by summing over all the columns of $D_i$, $\hat{y_i} = \text{colsum}(D_i)$. This can be equivalently expressed as $\hat{y_i} = D_i \mathbf{1}$, where $\mathbf{1}$ is a $d$-dimensional vector of all ones.

\begin{figure}[h]
\centerline{\includegraphics[width = 0.49\textwidth]{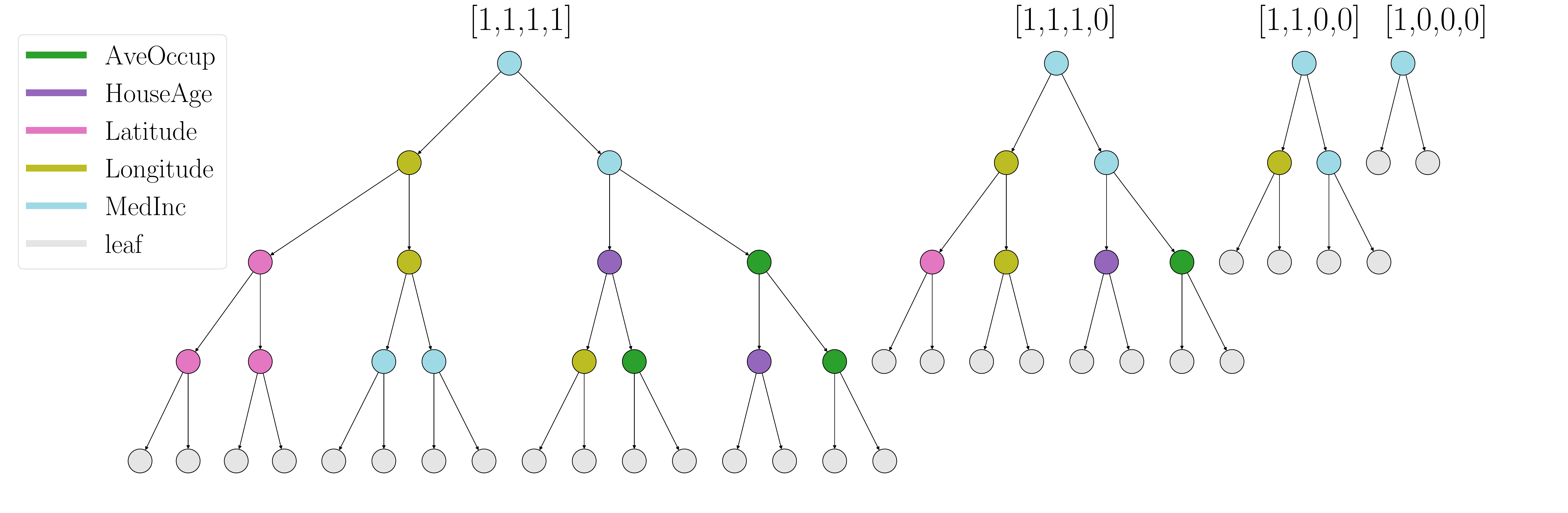}}
\caption{Pruning a decision tree with vector $z_i$.}
\label{pruneviz.fig}
\end{figure}

Consider the case where we want to prune the deepest layer of tree $T_i$. The predictions of this new tree, of depth $d-1$, can be obtained by summing over all but the last column of $D_i$.  Let $z_i$ be a vector in $\{0,1\}^d$ with $d-1$ ones followed by a single zero, $z_i = [ 1, 1, \ldots, 1, 0]$. The predictions of the pruned tree equal $D_i z_i$. To prune tree $T_i$ to depth $d - k$, set $z_i \in \{0,1\}^{d}$ as a vector of $d -k $ ones followed by $k$ zeros. Figure \ref{pruneviz.fig} visualizes this effect on a depth 4 decision tree. Note that setting $z_i = \mathbf{0}$ removes tree $T_i$ from the ensemble. 

\noindent{\bf Ensemble Pruning } Given tree ensemble $T_i,\ i \in [n]$, compute depth-difference matrices $D_i, \ i \in [n]$ for each tree. The predictions of the original ensemble can be represented by $\hat{y} = \sum_{i=1}^n \gamma D_i z_i$, where $z_i = \textbf{1} \mkern9mu \forall \ i \in [n]$. To prune the ensemble, modify the $z_i$ vectors. For example, setting $z_1 = [1,0,\ldots, 0]$ prunes the first tree in the ensemble to depth 1 and setting $z_2 = \mathbf{0}$ removes the second tree entirely.

\subsection{Ensemble Optimization Problem}
Notationally, let $(z_i)_k$ represent the $k$-th element of vector $z_i$. \textsc{ForestPrune} uses the following optimization problem to prune depth layers from an ensemble:
\begin{mini!}|s|
{z_1, \ldots, z_n} {\frac{1}{m} \Vert y - \gamma \sum_{i}^n D_i z_i\Vert_2^2 + \frac{\alpha}{K} \sum_{i=1}^n \Vert W_i z_i \Vert_1 \label{opt1obj}}{\label{optimizationproblem}}{}
\addConstraint{(z_i)_k}{\in \{0,1\} \quad  \forall \ i \in [n] , \ k \in [d] \label{opt1c1}}
\addConstraint{(z_i)_{k_1}}{\geq (z_i)_{k_2} \quad  \forall \ i \in [n] ,\ k_1 < k_2 \label{opt1c2}}
\end{mini!}

where $W_i \in \mathbb{R}^{d \times d}$ is a diagonal weight matrix assigned to each tree $T_i$, of the form $ W_{i} = \text{diag}(w_{i,1}, \ldots, w_{i,d})$. These weights are prespecified, and we discuss weighting schemes in \S\ref{weightschemes}. A normalization constant $K$ is computed from $W_i, \ i \in [n]$, and $\alpha$  is the regularization parameter.

Each decision variable $(z_i)_k$ is binary (\ref{opt1c1}) and represents whether to include the $k$-th layer of tree $T_i$ in the processed ensemble. Each tree $T_i$ has a corresponding decision vector $z_i \in \{0,1\}^{d}$ that contains $d$ decision variables. The goal of \textsc{ForestPrune} is to minimize regularized loss (\ref{opt1obj}) with respect to decision vectors $z_i, \ i \in [n]$. Constraint \eqref{opt1c2} ensures that the pruned trees are contiguous since trees should not be able to skip depth layers. For example, a solution $z_i = [ 1, 0, 1, \ldots , 1, 0 ]$ is infeasible since a tree cannot skip depth layer 2 and proceed to depth layer 3. Trees can only be pruned from the bottom up. 

Consider the regularized loss function in objective (\ref{opt1obj}). The first term of the function is least-squares loss and the second term is the regularization penalty. The parameter $\alpha$ controls regularization; larger values of $\alpha$ encourage shallower trees in the processed ensemble. Each entry of $W_i$, $w_{i,k}$ for $k \in [d]$, is nonnegative and represents the weight associated with including depth layer $k$ of tree $T_i$ into the processed ensemble. Since the entries of $W_i$ and $z_i$ are both nonnegative, the term $\Vert W_i z_i \Vert_1$ is equivalent to summing the elements of vector $W_i z_i$. Finally, $K$ is a normalization constant set such that $K = \sum_{i=1}^n \sum_{k = 1}^d w_{i,k}$---this ensures that $\alpha$ remains within a reasonable range which facilitates  tuning the regularization parameter.

\subsection{Weighting Schemes}\label{weightschemes}

We specify weight matrices $W_i, \ i\in [n]$ to produce post-processed ensembles with different parsimony properties. While various choices of $W$ are possible, we discuss a couple of choices that we explored in detail. In \textbf{depth-weighting}, we assign $w_{i,k} = 1$ if the depth of decision tree $T_i$ is less than or equal to $k$, otherwise we assign $w_{i,k} = 0$. We set normalization constant $K = nd$. This directly penalizes the total number of layers in the ensemble and produces ensembles with fewer trees. In \textbf{node-weighting}, we assign $w_{i,k}$ equal to the number of nodes in layer $k$ of tree $i$. The normalization constant $K$ is equal to the total number of nodes in the ensemble. This directly penalizes ensemble size and produces the most compact processed ensemble, with the shallowest trees.
\begin{figure}[h!]
  \begin{minipage}[c]{0.34\textwidth}
    \includegraphics[width=\textwidth]{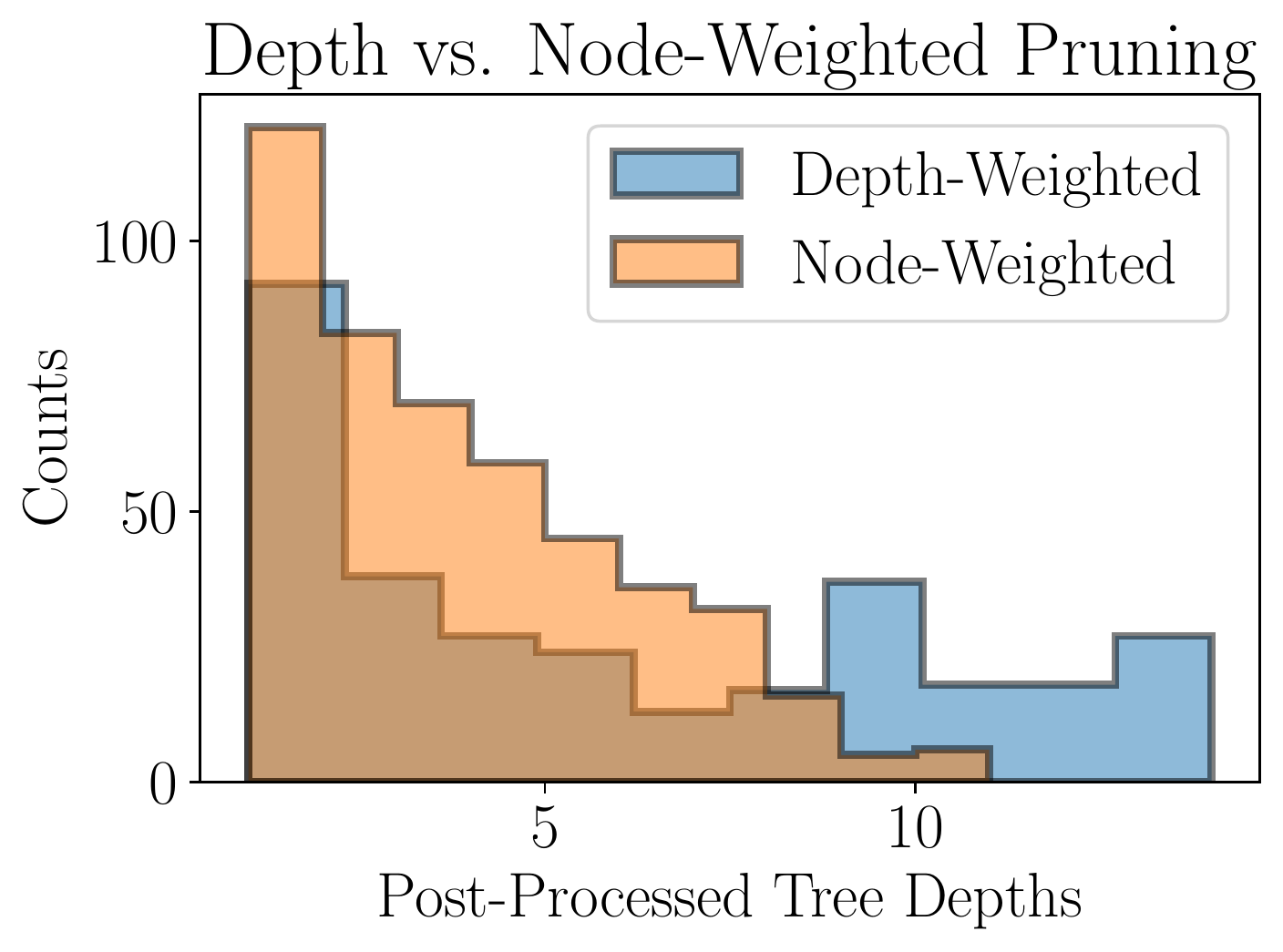}
  \end{minipage}\hfill
  \begin{minipage}[c]{0.14\textwidth}
    \caption{Node-weighting encourages shallow trees in the pruned ensemble.} \label{node_v_depth.fig}
  \end{minipage}
\end{figure}
Figure~\ref{node_v_depth.fig} compares the effect of depth vs. node-weighted \textsc{ForestPrune}. Node-weighting produces shallower trees compared to depth-weighting. In addition, the total size of the node-weighted pruned ensemble is approximately three times smaller than the depth-weighted one, \textbf{7247} vs. \textbf{21588} nodes. However, the ensemble processed by node-weighted \textsc{ForestPrune} contains more trees than the depth-weighted one, \textbf{95} vs. \textbf{62} trees. The test errors of the post-processed ensembles are comparable under both weighting schemes. Empirically, we observe that node-weighting encourages shallower trees and more compact ensembles, while depth-weighting produces an ensemble with fewer trees.

\subsection{Polishing Schemes} \label{polish.section}

Solving Problem \eqref{optimizationproblem} yields solution vectors $z_i^*, \ i \in [n]$. These vectors, combined with depth-difference matrices $D_i, \ i \in [n]$ represent the pruned ensemble, with trees $T_i^*(x) = D_i z_i^*, \ i \in [n]$. We can adjust the weights of trees in this pruned ensemble to better fit the model to $y$. This especially helps for boosting, where trees are grown sequentially, since pruning a tree breaks the boosting sequence. We can also select a small subset pruned trees for further analysis by setting the weights of some trees to 0. This may be especially useful for bagging, where the pruned ensemble typically consists of many shallow trees. Reducing the number of trees improves the interpretability of the final model.

Motivated by the above discussion, our polishing scheme reweights the post-processed ensemble via the following optimization criterion:
\begin{mini}|s|
{\beta_1 \ldots \beta_n}{ \frac{1}{m} \Vert y - \gamma\sum_{i=1}^n \beta_i D_i z_i^*\Vert_2^2 + \alpha_2 \Vert \beta \Vert_{\rho}^\rho }{\label{ridgepolishproblem}}{}
\end{mini} where coefficients $\beta_i, \ i \in [n]$ are the weights assigned to each tree, $\Vert \beta \Vert_{\rho}^{\rho}$ is the regularization penalty, and $\alpha_2$ is the regularization hyperparameter. We set $\rho = 2$ for \textbf{ridge polishing} to reweight the trees. The ridge penalty ensures a unique estimator in  $\beta$ when $n > m$ and offers stability when the bases elements $D_i z_i^*$ are correlated; we observe empirically that small values of $\alpha_2 \approx 10^{-2}$ work well. We set $\rho = 0$ for \textbf{best subset polishing} and vary $\alpha_2$ to select a small subset of trees for further analysis. To find good solutions for $\beta$, we use iterative hard thresholding \citep{blumensath2009iterative} or mixed-integer programming solvers \citep{bertsimas2016best}. In \S\ref{joint_opt.section}, we discuss a variant of \textsc{ForestPrune} combines polishing with pruning by jointly optimizing $\beta$ and $z$.

\subsection{Putting Together the Pieces}

\begin{algorithm}[h]
\scriptsize 
\caption{\textsc{ForestPrune} Framework}
\label{frameworkalgo}
\DontPrintSemicolon
  
  \KwInput{$T_i, \ i \in [n]$,  $X \in \mathbb{R}^{m \times p}$, $y \in \mathbb{R}^m$, $\alpha$}
  
  Compute $D_i, \ i \in [n]$.
  
  Choose weighting scheme $W_i, \ i \in [n]$.
  
  Solve Problem \eqref{optimizationproblem} for $z_i^*, \ i \in [n]$.
  
  \uIf{polish = True}{
  Solve Problem \eqref{ridgepolishproblem} for $\beta_i^*, \ i \in [n]$.
  }
  \Else{
  $\beta^* = \textbf{1}$.
  }
  \KwOutput{$z^*,\beta^*$}
\end{algorithm}

Algorithm \ref{frameworkalgo} presents the overall framework of \textsc{ForestPrune}. We recommend using both node-weighting and the polishing heuristic to compactify ensembles.

Ensemble optimization Problem \eqref{optimizationproblem} can be expressed as a mixed-integer optimization problem \citep{wolsey1999integer}, which can be computationally challenging for large problems \citep{bertsimas2016best,bertsimas2020sparse,hazimeh2021sparse}. Here we propose a novel algorithm based on block coordinate descent to obtain high quality solutions for Problem \eqref{optimizationproblem}, that appear to work well for the problem-scales we study.

\section{OPTIMIZATION ALGORITHM}

Note that Problem \eqref{optimizationproblem} can be written as:
\begin{mini!}|s|
{z_1 \ldots z_n}{L(z_1 \ldots z_n) + \sum_{i=1}^{n}g_i(z_i) \label{opt3obj} }{\label{optimizationproblem3}}{}
\addConstraint{z_i}{\in C_i \quad \forall \ i \in [n] \label{opt3c1} }
\end{mini!}
where $L(z_1 \ldots z_n) = (1/m) \Vert y - \gamma \sum_{i}^n D_i z_i\Vert_2^2$
is a smooth function and 
$\sum_{i=1}^n g_i(z_i) = (\alpha/K)\sum_{i=1}^n \Vert W_i z_i \Vert_1$ is separable across blocks $z_i$'s. Constraints \eqref{opt1c1} and \eqref{opt1c2} can be represented by constraint \eqref{opt3c1}, where $C_i$ is the set of vectors in $\{0,1\}^d$ that satisfy condition $(z_i)_{k_1} \geq (z_i)_{k_2} \mkern9mu \forall \ z_i \in C_i, \ k_1 < k_2$. 

The non-smooth part of the objective in \eqref{opt3obj} and constraint \eqref{opt3c1} are both separable across $z_i$'s. Motivated by the success of cyclic block coordinate descent methods (CBCD) for large-scale sparse regression problems \citep{wright2015coordinate,l0learn}, we apply CBCD methods to Problem \eqref{optimizationproblem3}.  CBCD yields a sequence of decreasing objective values, and since objective \eqref{opt3obj} is nonnegative the sequence will converge. 

\subsection{Cyclic Block Coordinate Descent (CBCD)}
\label{CBCD.section}
Consider a partition of the decision variables in Problem \eqref{optimizationproblem3} into the blocks $z_i, \ i \in [n]$ with $z_i \in C_i$. CBCD works as follows. Initialize the algorithm by setting all blocks equal to the zero vector, $z_i = \textbf{0} \mkern9mu  \forall \ i \in [n]$, and set index $\omega = 1$. Start with the first block of decision variables $z_{\omega} = z_1$ and minimize the objective with respect to $z_\omega$ while holding the other blocks $\{z_2 \ldots z_n\}$ constant. Repeat this procedure, cycling through the blocks by incrementing $\omega$, until the objective value converges.

\noindent{\bf Block Update } Given fixed index $\omega$, let set $\delta = \{ 1 \ldots n\} \setminus \omega$. To update block $z_{\omega}$, we solve the following problem:
\begin{mini!}|s|
{z_{\omega}}{\frac{1}{m} \Vert \tilde{y}_{\delta} - \gamma D_{\omega}z_\omega \Vert_2^2 + \frac{\alpha}{K} \sum_{k=1}^{d} (W_{\omega} z_{\omega})_k \label{opt4obj}}{\label{optimizationproblem4}}{}
\addConstraint{(z_{\omega})_k}{ \in \{0,1\}   \quad \forall \ k \in [d]\label{opt4c1}}
\addConstraint{(z_{\omega})_{k_1}}{ \geq (z_{\omega})_{k_2}  \quad \forall \ k_1 < k_2, \label{opt4c2}}
\end{mini!}

where $\tilde{y}_{\delta} = y - \gamma\sum_{i \in \delta} D_i z_i$. This is equivalent to solving Problem \eqref{optimizationproblem} or \eqref{optimizationproblem3} with respect to $z_{\omega}$ while holding all the other blocks constant. As we show below Problem \eqref{optimizationproblem4}, despite its non-convexity, can be solved to optimality.

In Problem \eqref{optimizationproblem4},  $z_{\omega}$ is constrained to be a contiguous binary vector where the zeros occur after the ones. As a result, $z_{\omega}$ can only attain $d+1$ unique possible values. For example if $d = 3$, the candidates for $z_{\omega}$ are $[1,1,1]$, $[1,1,0]$, $[1,0,0]$, and $[0,0,0]$. By enumerating through all possible candidates for $z_{\omega}$, Problem \eqref{optimizationproblem4} can be solved very quickly.

Each block update requires a single pass through $d+1$ candidates, where the objective is evaluated for each candidate. Evaluating the objective requires approximately $m \cdot  d$ flops, so each block update requires approximately $m \cdot  d^{2}$ flops. Typically the maximal depth $d$ is below 20, so when $m >> d$ the per-iteration complexity of CBCD is linear in the number of training samples, $O(m)$.

\noindent{\bf Local Search Heuristic }
 Due to the non-convexity of Problem \eqref{optimizationproblem3}, CBCD can get stuck in local suboptimal solutions. To improve solution quality, we employ a local combinatorial search procedure; a similar procedure was used in L0Learn in the context of sparse high-dimensional regression \citep{hazimeh2021sparse}. 
 
 After CBCD converges to a local solution $\hat{z}_i, \ i \in [n]$, partition the solution into two sets. Let $S$ denote the set of indices for which $\hat{z}_i \neq \mathbf{0}$ and let $S^c$ denote its complement. Select a random index $\xi \in S$ and  set $z_{\xi} = \mathbf{0}$. Find the index $i^* \in S^c$ for which $D_i\hat{z}_{i^*}$ is the most correlated with the response $y$ and set $z_{i^*} = \mathbf{1}$. Run CBCD with this new candidate solution till convergence and continue interlacing CBCD with local search steps, until local search no longer improves the objective value. 
 
 We observe that the local search heuristic discussed above is effective in guiding CBCD methods from suboptimal local solutions. A drawback of this procedure is that it can be expensive to find the best swap-coordinate $i^* \in S^c$ that would result in the largest improvement in the current objective. In fact, this would require computing the correlation coefficient between each block in $S^c$ and $y$. We present a heuristic that we found to work quite well for our problem in the context of boosting: we always select the block in $S^c$ with the smallest index, let $i^* = \min  \{i \ | \ \hat{z}_i \in S^c\}$. In boosting ensembles, trees grown early in the boosting sequence are generally more correlated with the response. We can also extend this rule to bagging ensembles by first sorting the trees in the ensemble by training performance. The smallest index rule ensures that the local search heuristic always swaps in the earliest-grown tree, for negligible computational cost. 

The complete CBCD algorithm with local search is presented in Algorithm \ref{cbcdalgo}.

\begin{algorithm}[h]
\scriptsize 
\caption{\textsc{ForestPrune} CBCD}
\label{cbcdalgo}
\DontPrintSemicolon
  \KwInput{ $D_i, \ W_i \ \text{for} \ i \in [n]$, $y \in \mathbb{R}^m$, $\alpha$}
  
  \textbf{Initialize} $z_i = \textbf{0} \mkern9mu \forall \ i \in [n]$, $\omega = 1$ 
  
  \Repeat{objective no longer improves}{
  
  \Repeat{converged}{
  
  $\omega = \omega \mod n$
  
  Solve Problem \eqref{optimizationproblem4} to update $z_\omega$.
  
  $\omega = \omega + 1$
  }
  
  Local search.
  }
  \KwOutput{$z_i, \ i \in [n]$}
\end{algorithm}

\subsection{Regularization Paths} \label{regularizationpath}

We use warm start continuation with Algorithm \ref{cbcdalgo} to efficiently compute the entire regularization path of  $z_i$'s, across tuning parameter $\alpha$. Start with a value of $\alpha$ sufficiently large such that $z_i^* = \textbf{0}, \mkern9mu \forall \ i \in [n]$ and decrement $\alpha$ using the previous solution as a warm start until the full model is reached \citep{friedman2010regularization}.  This provides a sequence of solutions with varying ensemble sizes that a practitioner can use to quickly select a suitable model. 

\begin{figure}[h]
\centerline{\includegraphics[width = 0.5\textwidth]{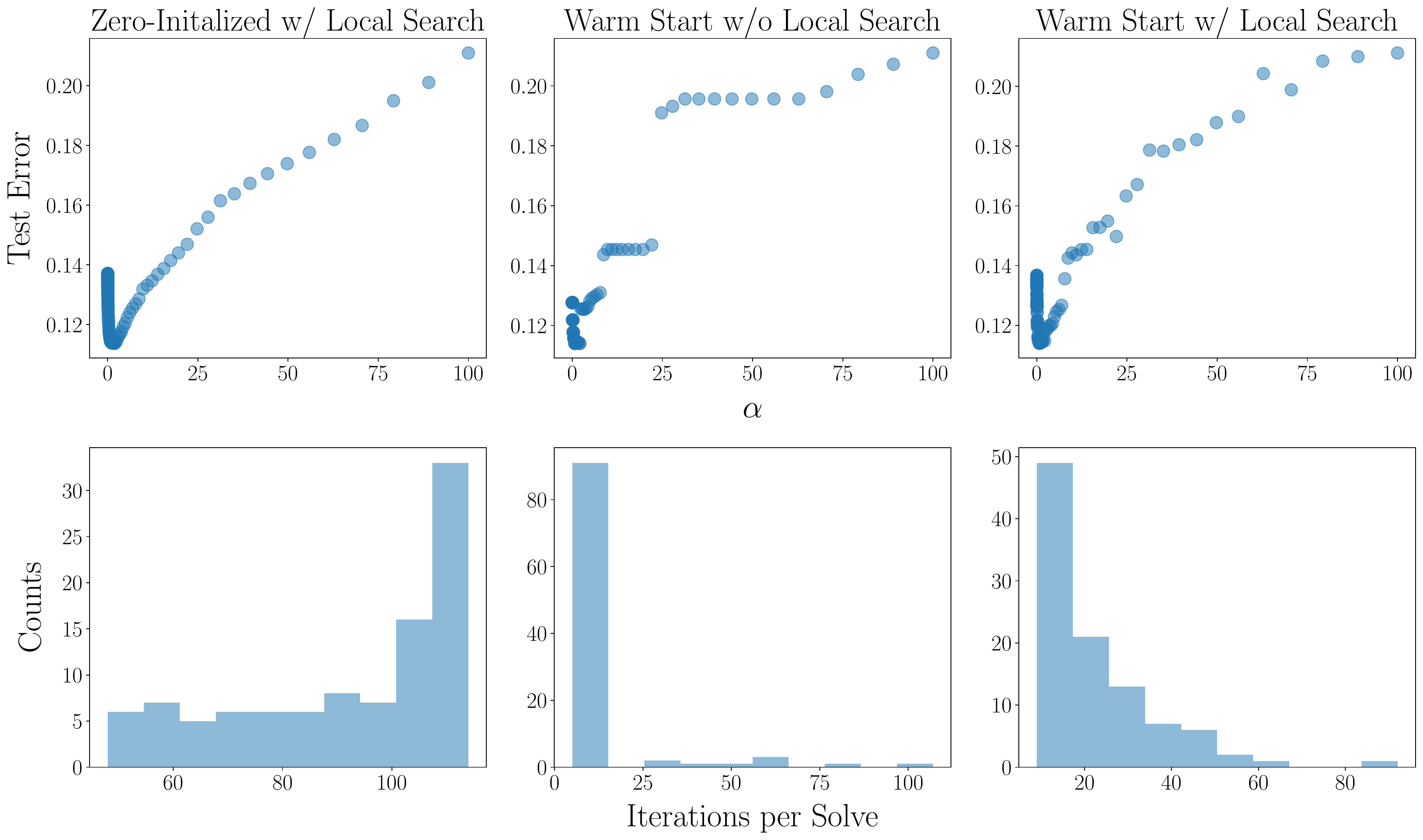}}
\caption{Warm start continuation with local search reduces the iterations (\# of passes over all features) required to compute the regularization path while avoiding local minima.}
\label{ws_paths.fig}
\end{figure}

When warm start continuation is used without local search, CBCD tends to get stuck at suboptimal local solutions. This can be seen in the middle plots of Figure \ref{ws_paths.fig}. In this example, CBCD with warm start continuation is used to compute the entire regularization path of \textsc{ForestPrune}. The scatter plot shows that the regularization path is nearly piecewise, and the histogram shows that CBCD often terminates immediately after initialization. Combining warm start continuation with local search resolves this issue, as seen in the rightmost plots of Figure \ref{ws_paths.fig}. The regularization path for CBCD with warm starts and local search is very similar to the regularization path when CBCD is always zero-initialized (leftmost plots), however, our warm start procedure requires substantially fewer iterations, i.e., the \# of passes over all features. As a result, this procedure makes $\alpha$ very efficient to tune


\subsection{Computation Time}

We now discuss the time required to compute good solutions to Problem \eqref{optimizationproblem} using our \textsc{ForestPrune} CBCD algorithm. The computation time of CBCD depends on the number of observations in the data and the number of trees (blocks) in the ensemble. We fit tree ensembles with 100, 500, and 1000 trees of depth 6 on datasets with 1145, 15437, and 50164 rows. We set regularization parameter $\alpha = 1$ and use \textsc{ForestPrune} CBCD to compute solutions to Problem \eqref{optimizationproblem}. Also, we use CBCD with warm start continuation and local search to compute the entire regularization path.

We compare CBCD against this benchmark. We relax integrality constraint \eqref{opt1c1} in Problem \eqref{optimizationproblem}; the relaxed problem is a convex second order conic program. We use the open-source solver ECOS \citep{domahidi2013ecos} and the proprietary industrial-grade solver MOSEK \citep{andersen2000mosek} to compute solutions to this relaxed problem. Table \ref{timingtable.fig} shows the results of this experiment  conducted on a personal laptop with a 2.80 GHz Intel Core i7 processor.

\textsc{ForestPrune} CBCD computes solutions to Problem \eqref{optimizationproblem} orders of magnitude faster than the time it takes ECOS or MOSEK to solve the relaxed problem. Problem \eqref{optimizationproblem} is difficult to solve for larger instances: For an ensemble of 100 trees of depth 6 fit on a dataset with 50164 rows, the corresponding optimization problem has 600 decision variables and even after relaxing the integrality constraints, it takes an industrial-grade solver over 30 minutes to reach a solution. In contrast, our specialized CBCD algorithm can find a good solution in seconds. \textsc{ForestPrune} CBCD can also efficiently compute the entire regularization path for $\alpha$, for example, for Problem \eqref{optimizationproblem} with 1145 rows and 100 trees, it takes CBCD 0.2 seconds to compute the entire regularization path. MOSEK takes over 11 minutes to compute the same regularization path for the relaxed problem. 

\begin{table}[]
\scalebox{0.48}{
\begin{tabular}{|cccc|l|cccc|}

\cline{1-4} \cline{6-9}
\multicolumn{4}{|c|}{\cellcolor[HTML]{C0C0C0}{\color[HTML]{333333} \textbf{ForestPrune CBCD Single Solve}}}                                                                                                                                                                                                                          &  & \multicolumn{4}{c|}{\cellcolor[HTML]{C0C0C0}\textbf{Linear Relaxation (ECOS) Single Solve}}                                                                                                                                                                                                                   \\ \cline{1-4} \cline{6-9} 
\multicolumn{1}{|c|}{\textbf{\begin{tabular}[c]{@{}c@{}}Rows / \\ Trees\end{tabular}}} & \multicolumn{1}{c|}{\textbf{100}}                                                  & \multicolumn{1}{c|}{\textbf{500}}                                                   & \textbf{1000}                                                    &  & \multicolumn{1}{c|}{\textbf{\begin{tabular}[c]{@{}c@{}}Rows / \\ Trees\end{tabular}}} & \multicolumn{1}{c|}{{\color[HTML]{333333} \textbf{100}}}                     & \multicolumn{1}{c|}{{\color[HTML]{333333} \textbf{500}}}                     & {\color[HTML]{333333} \textbf{1000}}                    \\ \cline{1-4} \cline{6-9} 
\multicolumn{1}{|c|}{\textbf{1145}}                                                    & \multicolumn{1}{c|}{\cellcolor[HTML]{FFFFFF}{\color[HTML]{333333} 0.1 $\pm$ 0.01}} & \multicolumn{1}{c|}{\cellcolor[HTML]{FFFFFF}{\color[HTML]{333333} 0.8 $\pm$ 0.10}}  & \cellcolor[HTML]{FFFFFF}{\color[HTML]{333333} 1.5 $\pm$ 0.16}    &  & \multicolumn{1}{c|}{\textbf{1145}}                                                    & \multicolumn{1}{c|}{{\color[HTML]{333333} 9.7 $\pm$ 0.20}}                   & \multicolumn{1}{c|}{{\color[HTML]{333333} 829.8 $\pm$ 26.72}}                & \cellcolor[HTML]{FFCCC9}{\color[HTML]{333333} $>$ 1800} \\ \cline{1-4} \cline{6-9} 
\multicolumn{1}{|c|}{\textbf{15437}}                                                   & \multicolumn{1}{c|}{\cellcolor[HTML]{FFFFFF}{\color[HTML]{333333} 0.2 $\pm$ 0.03}} & \multicolumn{1}{c|}{\cellcolor[HTML]{FFFFFF}{\color[HTML]{333333} 0.9 $\pm$ 0.10}}  & \cellcolor[HTML]{FFFFFF}{\color[HTML]{333333} 2.2 $\pm$ 0.25}    &  & \multicolumn{1}{c|}{\textbf{15437}}                                                   & \multicolumn{1}{c|}{{\color[HTML]{333333} 184.3 $\pm$ 0.60}}                 & \multicolumn{1}{c|}{\cellcolor[HTML]{FFCCC9}{\color[HTML]{333333} $>$ 1800}} & \cellcolor[HTML]{FFCCC9}{\color[HTML]{333333} $>$ 1800} \\ \cline{1-4} \cline{6-9} 
\multicolumn{1}{|c|}{\textbf{50164}}                                                   & \multicolumn{1}{c|}{\cellcolor[HTML]{FFFFFF}{\color[HTML]{333333} 3.0 $\pm$ 0.02}} & \multicolumn{1}{c|}{\cellcolor[HTML]{FFFFFF}{\color[HTML]{333333} 69.7 $\pm$ 2.60}} & \cellcolor[HTML]{FFFFFF}{\color[HTML]{333333} 131.0 $\pm$ 20.03} &  & \multicolumn{1}{c|}{\textbf{50164}}                                                   & \multicolumn{1}{c|}{\cellcolor[HTML]{FFCCC9}{\color[HTML]{333333} $>$ 1800}} & \multicolumn{1}{c|}{\cellcolor[HTML]{FFCCC9}{\color[HTML]{333333} $>$ 1800}} & \cellcolor[HTML]{FFCCC9}{\color[HTML]{333333} $>$ 1800} \\ \cline{1-4} \cline{6-9} 
\multicolumn{4}{|c|}{\cellcolor[HTML]{C0C0C0}\textbf{ForestPrune CBCD Regularization Path}}                                                                                                                                                                                                                                          &  & \multicolumn{4}{c|}{\cellcolor[HTML]{C0C0C0}\textbf{Linear Relaxation (MOSEK) Single Solve}}                                                                                                                                                                                                                               \\ \cline{1-4} \cline{6-9} 
\multicolumn{1}{|l|}{\textbf{\begin{tabular}[c]{@{}l@{}}Rows /\\  Trees\end{tabular}}} & \multicolumn{1}{c|}{{\color[HTML]{333333} \textbf{100}}}                           & \multicolumn{1}{c|}{{\color[HTML]{333333} \textbf{500}}}                            & {\color[HTML]{333333} \textbf{1000}}                             &  & \multicolumn{1}{c|}{\textbf{\begin{tabular}[c]{@{}c@{}}Rows /\\  Trees\end{tabular}}} & \multicolumn{1}{c|}{{\color[HTML]{333333} \textbf{100}}}                     & \multicolumn{1}{c|}{{\color[HTML]{333333} \textbf{500}}}                     & {\color[HTML]{333333} \textbf{1000}}                    \\ \cline{1-4} \cline{6-9} 
\multicolumn{1}{|c|}{\textbf{1145}}                                                    & \multicolumn{1}{l|}{0.2 $\pm$  0.01}                                               & \multicolumn{1}{l|}{1.5$\pm$ 0.01}                                                   & \multicolumn{1}{l|}{4.3$\pm$ 0.10}                                &  & \multicolumn{1}{c|}{\textbf{1145}}                                                    & \multicolumn{1}{c|}{{\color[HTML]{333333} 7.0 $\pm$ 0.28}}                   & \multicolumn{1}{c|}{{\color[HTML]{333333} 21.4 $\pm$ 0.27}}                  & {\color[HTML]{333333} 59.8 $\pm$ 1.06}                  \\ \cline{1-4} \cline{6-9} 
\multicolumn{1}{|c|}{\textbf{15437}}                                                   & \multicolumn{1}{l|}{2.3$\pm$ 0.08}                                                  & \multicolumn{1}{l|}{11.6$\pm$ 0.15}                                                  & \multicolumn{1}{l|}{24.2$\pm$ 0.45}                               &  & \multicolumn{1}{c|}{\textbf{15437}}                                                   & \multicolumn{1}{c|}{{\color[HTML]{333333} 140.5 $\pm$ 2.04}}                 & \multicolumn{1}{c|}{{\color[HTML]{333333} 874.4 $\pm$ 59.93}}                & \cellcolor[HTML]{FFCCC9}{\color[HTML]{333333} $>$ 1800} \\ \cline{1-4} \cline{6-9} 
\multicolumn{1}{|c|}{\textbf{50164}}& \multicolumn{1}{l|}{10.2 $\pm$ 0.10}                                                 & \multicolumn{1}{l|}{106.3$\pm$ 1.20}                                                 & \multicolumn{1}{l|}{206.5$\pm$ 1.23}                              &  & \multicolumn{1}{c|}{\textbf{50164}}                                                   & \multicolumn{1}{c|}{\cellcolor[HTML]{FFCCC9}{\color[HTML]{333333} $>$ 1800}} & \multicolumn{1}{c|}{\cellcolor[HTML]{FFCCC9}{\color[HTML]{333333} $>$ 1800}} & \cellcolor[HTML]{FFCCC9}{\color[HTML]{333333} $>$ 1800} \\ \cline{1-4} \cline{6-9} \end{tabular}}
\caption{Timing results in seconds. The red cells indicate that the method did not converge within 30 minutes.}
\label{timingtable.fig}
\end{table}


\subsection{Joint Optimization for Pruning and Polishing }
\label{joint_opt.section}

As mentioned in \S\ref{polish.section}, we present a variant of \textsc{ForestPrune} that combines polishing with pruning by jointly optimizing $z$ and $\beta$. The corresponding optimization problem considers the following:
\begin{mini}|s|
{z, \beta} {\frac{1}{m} \Vert y - \gamma \sum_{i}^n \beta_i D_i z_i\Vert_2^2 \breakObjective{+ \frac{\alpha}{K} \sum_{i=1}^n \Vert W_i z_i \Vert_1  + \alpha_2 \Vert \beta \Vert_{\rho}^\rho, }\label{optjointobj}}{}{}
\end{mini} 
and inherits the constraints from Problem \eqref{optimizationproblem}. Variable $\beta$ separates across blocks so Problem \eqref{optjointobj} is again block separable. 

We can apply our CBCD algorithm to this problem with a slight modification to how we conduct block updates. Each block update now considers an optimization problem of the form:
\begin{mini}|s|
{z_{\omega}, \beta_{\omega}}{\frac{1}{m} \Vert \tilde{y}_{\delta} - (\gamma  D_{\omega}z_\omega) \beta_{\omega} \Vert_2^2 \breakObjective{ + \frac{\alpha}{K} \sum_{k=1}^{d} (W_{\omega} z_{\omega})_k + \alpha_2 \Vert \beta_\omega \Vert_{\rho}^{\rho},}}{\label{optimizationproblemj1}}{}
\end{mini}
where $\tilde{y}_{\delta} = y - \gamma\sum_{i \in \delta} \beta_i D_i z_i$.  This new block update problem also inherits the constraints from Problem \eqref{optimizationproblem4}, and, as discussed in \S\ref{CBCD.section}, vector $z_w$ has $d+1$ unique possible values due to these constraints. Therefore, vector $q_{\omega} = \gamma  D_{\omega}z_\omega$ has $d+1$ unique possible values as well. For each unique value of $q_\omega$, we solve this problem for scalar $\beta_{\omega}$:
\begin{mini}|s|
{\beta_{\omega}}{\frac{1}{m} \Vert \tilde{y}_{\delta} -  \beta_{\omega} q_\omega \Vert_2^2 + \alpha_2 \Vert \beta_\omega \Vert_{\rho}^{\rho}.}{\label{optimizationproblemj2}}{}
\end{mini}
For ridge polishing ($\rho = 2$) this univariate ridge problem has a closed-form solution for $\beta_\omega$. For best subset polishing ($\rho = 0$) good solutions for $\beta_\omega$ can be obtained through hard thresholding \citep{hazimeh2020fast}. We return the $z_\omega$, $\beta_\omega$ pair that yields the lowest objective value for Problem \eqref{optimizationproblemj1} to complete the block update. The other steps of our CBCD algorithm remain the same.

In practice, we observe that the ensembles post-processed by jointly optimizing $\beta$ and $z$ in \textsc{ForestPrune} can outperform the ensembles extracted by pruning followed by polishing. However, this improved performance comes at a computational cost, since our CBCD algorithm requires more iterations to converge over both sets of variables. We present this joint optimization method as an additional flexibility in our \textsc{ForestPrune} framework. Users can choose between joint optimization and pruning then polishing depending on their problem size and desired application.

\section{EXPERIMENTS}
We evaluate \textsc{ForestPrune} against these competing ensemble post-processing algorithms.

\noindent{\bf Baseline } A simple but useful baseline is to conduct fewer bagging/boosting iterations. For bagging, since trees are grown independently, we start with the full model and repeatedly remove randomly selected trees until a sparse model is obtained. For boosting, we start with the full boosting sequence and trim trees from the tail of the sequence.

\noindent{\bf Cost-Complexity Pruning} 
Cost-complexity pruning (CCP) prunes individual decision trees by recursively removing weak branches \citep{breiman2017classification}. Each decision tree is pruned independently of the other trees in the ensemble. To post-process the entire tree ensemble, we set the cost-complexity sparsity parameter to be the same for all trees. By varying this sparsity parameter, we control the trade-off between the degree of individual tree pruning and ensemble performance. 

\noindent{\bf LASSO }
Given a tree ensemble 
$T_i, i \in [n]$, LASSO ensemble pruning solves the following problem:
\begin{mini}|s|
{\beta_1 \ldots \beta_n}{ \frac{1}{m} \Vert y - \gamma\sum_{i}^n \beta_i T_i(X)\Vert_2^2 + \lambda \sum_{i=1}^n|\beta_i|,}{\label{lassoobjective}}{}
\end{mini}
where tree $i$ has prediction $T_i(X)$ and coefficient $\beta_i$. The $\ell_1$-penalty over the coefficients encourages sparsity.

\noindent{\bf Best Subset Tree Selection }
Best subset tree selection (BSTS) uses mixed integer quadratic programming (MIQP) to select the optimal subset of trees from an ensemble, subject to a size constraint. Let $n_i$ be the number of nodes used by tree $T_i$ and let $\nu$ be the maximum number of nodes to include in the pruned ensemble. The optimization problem can be expressed as follows,
\begin{mini!}|s|
{\beta_1 \ldots \beta_n}{ \frac{1}{m} \Vert y - \gamma\sum_{i}^n \beta_i T_i(X)\Vert_2^2}{\label{bssobjective}}{}
\addConstraint{(\beta_i, \: 1 - \zeta_i) \quad \text{SOS-1} \quad \forall \: i \in [n] \label{bssc1}}
\addConstraint{ \sum_{i=1}^n n_i \zeta_i \leq \nu \label{bssc2}}
\addConstraint{\zeta_i \in \{0,1\} \quad  \forall \ i \in [n], \label{bssc3}}
\end{mini!}
where constraint \eqref{bssc1} is a Type 1 Special Ordered Set (SOS-1) constraint \citep{bertsimas2016best} that ensures that $\zeta_i = 1$ if tree $T_i$ is assigned a nonzero weight. We implement BSTS in Gurobi \citep{gurobi} and warm start the solver using LASSO solutions. For smaller values of $\nu$, we are able to solve the MIQP to optimality in minutes.

\noindent{\bf Discussion on the Choice of Competing Methods }
There is a large corpus of work on selecting good subsets of trees from tree ensembles. Notable examples include a forward step-wise selection algorithm proposed by \cite{caruana2004ensemble}, a ranking algorithm prescribed by \cite{rokach2006selective}, and an evolutionary optimization algorithm proposed by \cite{qian2015pareto}. Most of these algorithms are heuristics to approximate the NP-hard problem of selecting the best subset of trees from an ensemble.

To simplify comparisons between \textsc{ForestPrune} and this family of competing algorithms we formulate BSTS, a new stronger benchmark that selects the optimal subset of trees using MIQP. Recent advancements in MIQP techniques \citep{bertsimas2016best, bertsimas2020sparse} have made this approach tractable for problems of the sizes that we are interested in. As such, we omit comparisons between \textsc{ForestPrune} and various heuristics in favor of evaluating our algorithm directly against our stronger benchmark. We include LASSO pruning as a competing algorithm since both LASSO and \textsc{ForestPrune} use warm start continuation to efficiently compute regularization paths~\citep{friedman2007pathwise}. We also compare \textsc{ForestPrune} against cost-complexity pruning for pruning bagging ensembles, since both algorithms can prune deep trees trees individually.

\subsection{Compact Bagging Ensembles} \label{baggingexperimet.section}

\noindent{\bf Procedure }
Bagging ensembles are generally robust to overfitting so pruning ensembles typically will not improve performance. To evaluate how well \textsc{ForestPrune} performs at compactifying bagging ensembles, we fix a threshold for acceptable performance loss and find the smallest pruned ensemble within this threshold. Repeat this procedure for 25 regression datasets in the OpenML repository \citep{OpenML2013} using 5-fold cross-validation; the full list of datasets can be found in the supplement. First, fit a bagging ensemble of 500 trees where each tree is grown to depth $d = 20$, and a subsample of $\sqrt{p}$ features is considered at each split. Next, fix a threshold for acceptable performance loss, $\phi \in \{0.01, 0.025, 0.05 \}$, and compute the entire regularization path for \textsc{ForestPrune}. We select $\alpha$ such that the validation loss of the pruned ensemble is within $\phi$ of the loss of the full model. Compare the reduction in ensemble size and the \% difference in test error between the pruned ensemble and the original. Finally, use the 4 competing methods described above to find the best model no larger than the ensemble pruned by \textsc{ForestPrune}, and compare the test performances of all methods.

 \begin{figure}[h]
\centerline{\includegraphics[width = 0.5\textwidth]{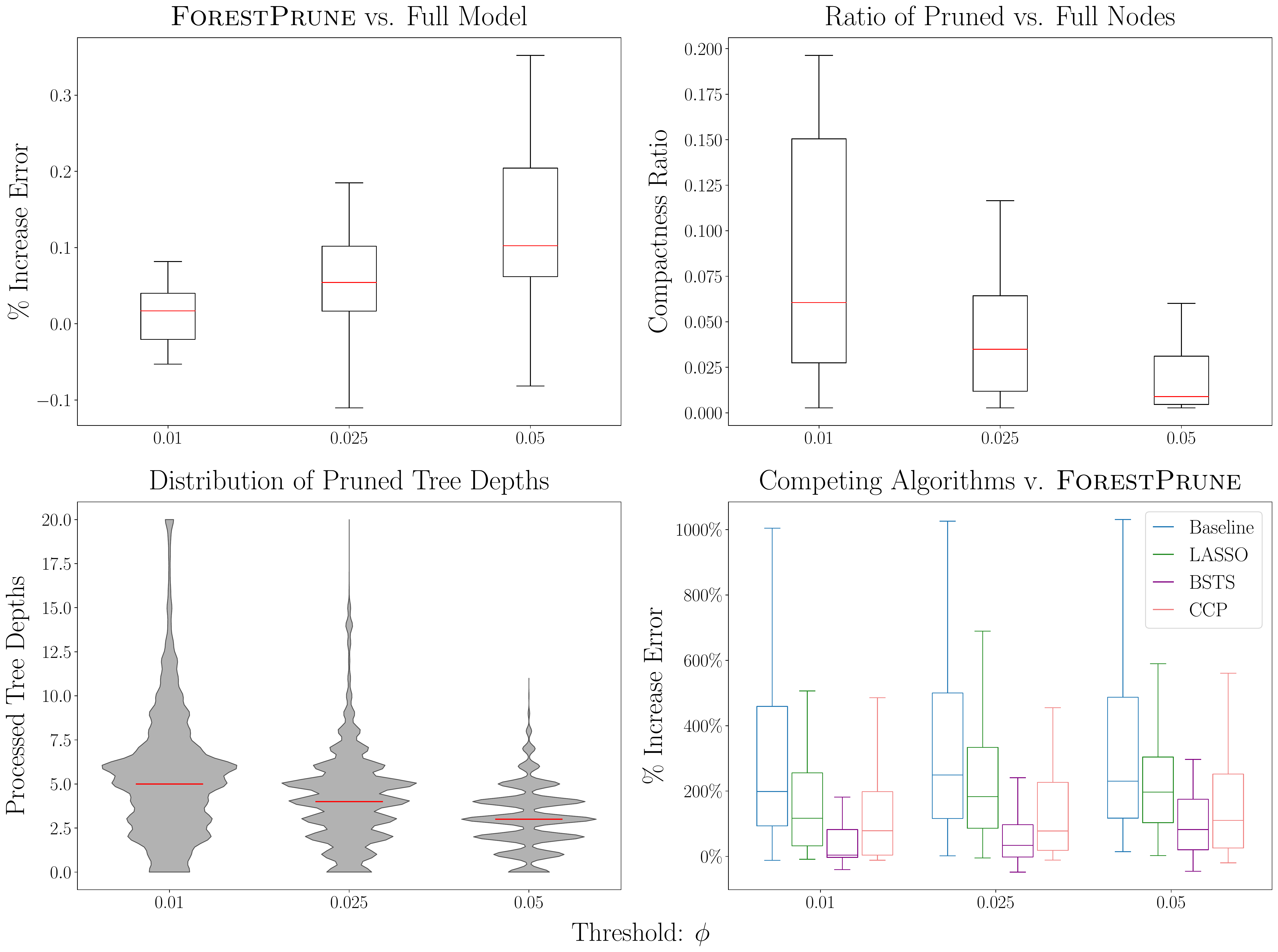}}
\caption{\textsc{ForestPrune} on bagging ensembles. \textsc{ForestPrune} can reduce ensemble size ~20x for a \textasciitilde 5\% increase in error.}
\label{baggingcompactresults.fig}
\end{figure} 
 
\noindent{\bf Results } Figure \ref{baggingcompactresults.fig} presents the results of the compact bagging experiment. The distributions are obtained across all folds and datasets in the experiment. The top left plot shows the \% increase in test error between the model pruned by \textsc{ForestPrune} and the full model; as the threshold for acceptable performance loss $\phi$ increases, the \% increase in test error rises correspondingly. The top right plot in Figure \ref{baggingcompactresults.fig} presents the compactness ratio, the ratio of the number of nodes in the pruned model over the number of nodes in the full model. The compactness ratio decreases as $\phi$ increases; if the threshold for acceptable performance loss is increased, the model can be made much smaller. The bottom left plot in Figure \ref{baggingcompactresults.fig} presents the distribution of tree depths in the pruned ensemble. Originally all trees in the bagging ensemble are grown to depth $d = 20$. Node-weighted \textsc{ForestPrune} produces much shallower trees, across all levels of $\phi$ the average tree depth in the pruned ensemble is around 5.

Across all levels of $\phi$, \textsc{ForestPrune} produces substantially smaller ensembles for modest increases in test error. For example, when $\phi = 0.01$, the ensembles post-processed by \textsc{ForestPrune} perform nearly the same as the full ensembles, with only a 1-5\% increase in test error, and are reduced 10-20 fold in size. The average depth of trees in the pruned ensemble is also reduced by a factor of 4. By varying the threshold $\phi$, \textsc{ForestPrune} can be tuned to balance performance and compactness.

\textsc{ForestPrune} also outperforms the competing algorithms. The bottom right plot in Figure \ref{baggingcompactresults.fig} shows the \% increase in test error between the model produced by \textsc{ForestPrune} and the models produced by the competing algorithms. The competing methods perform worse than \textsc{ForestPrune} if the \% increase in test error is positive.
Across all values of $\phi$, the distributions of the \% increase in test error between the competing algorithms and \textsc{ForestPrune} are almost entirely positive, with medians of  \{$\mathbf{220\%}$, $\mathbf{250\%}$, $\mathbf{250\%}$\} for the baseline, \{$\mathbf{130\%}$, $\mathbf{190\%}$, $\mathbf{220\%}$\} for LASSO, \{$\mathbf{5\%}$, $\mathbf{40\%}$, $\mathbf{80\%}$\} for BSTS, and \{$\mathbf{78\%}$, $\mathbf{78\%}$, $\mathbf{110\%}$\} for CCP.

The baseline, LASSO, and BSTS competing methods are only capable of excluding or selecting trees, and the size of a single deep tree may be larger than the shallow ensemble produced by \textsc{ForestPrune}. As a result, these post-processing methods are unable to produce a model similar in size to \textsc{ForestPrune} with comparable test performance. CCP, on the other hand, prunes individual trees, but prunes each tree independently from the other trees in the ensemble. \textsc{ForestPrune} outperforms CCP since \textsc{ForestPrune} prunes trees with respect to the overall performance of the ensemble.

\subsection{Compact Boosting Ensembles}

\noindent{\bf Procedure }
To evaluate how well \textsc{ForestPrune} compactifies and regularizes boosting ensembles, we evaluate the performance of the algorithm across the regularization path for $\alpha$. On the same datasets and folds described in \S\ref{baggingexperimet.section}, we fit stochastic gradient boosting ensembles of 250 depth 5 trees with $\gamma = 0.1$, subsampling 25\% of the training data for each tree. We post-process the ensembles and compare errors along the regularization paths of \textsc{ForestPrune} and the LASSO and baseline competing methods.

Computing regularization paths for BSTS is infeasible since the procedure requires repeatedly solving MIQP problems of increasing difficulty. To evaluate BSTS against \textsc{ForestPrune}, we fix an ensemble size budget and find the model selected by BSTS constrained by this budget. We use the regularization paths computed above to find the best models selected by \textsc{ForestPrune}, LASSO, and baseline within this budget and compare the test performances of all models (size budget boosting experiment).

\begin{figure}[h]
\centerline{\includegraphics[width = 0.5\textwidth]{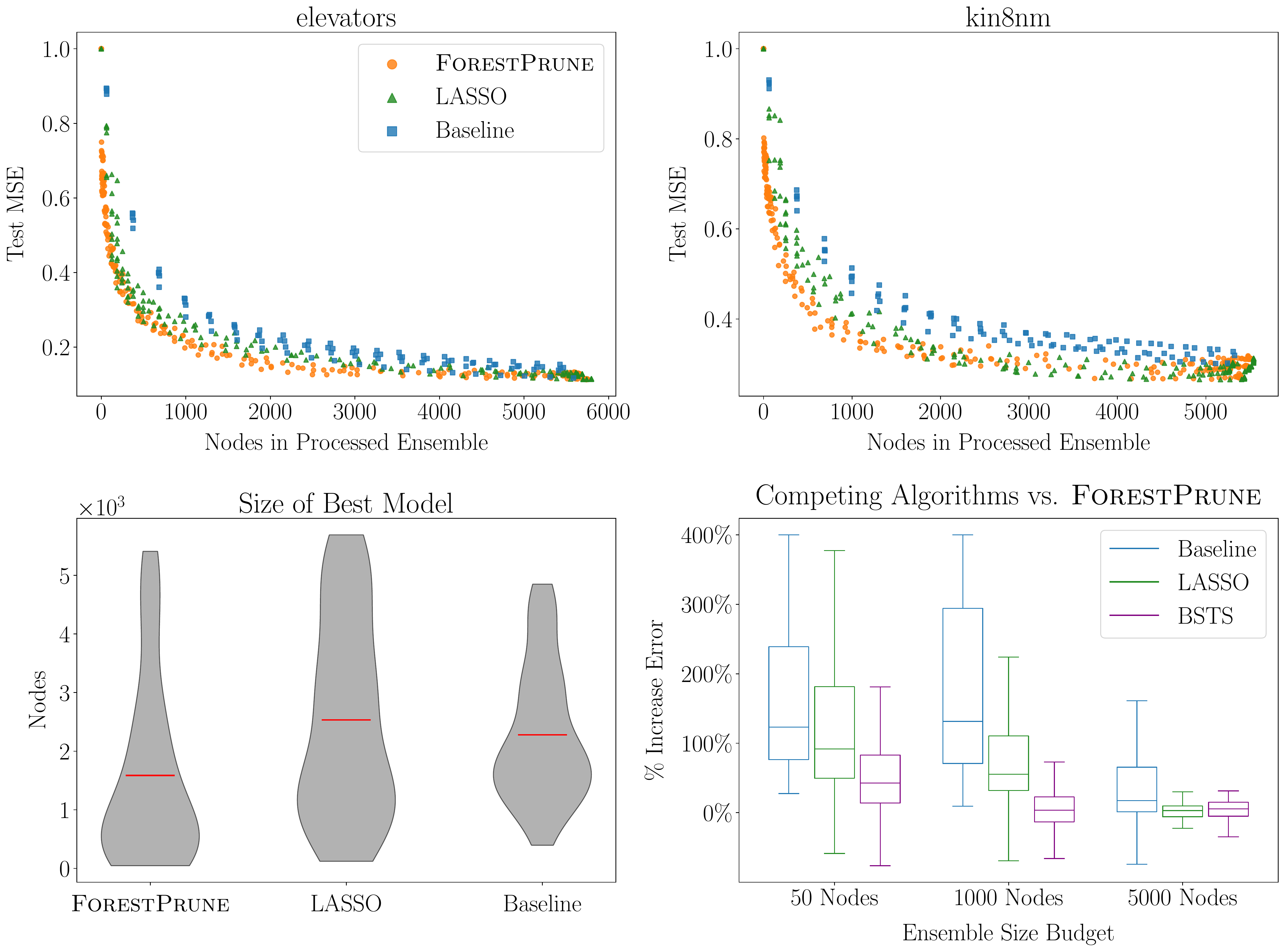}}
\caption{\textsc{ForestPrune} on boosting ensembles.}
\label{boostingcompactresults.fig}
\end{figure}

\noindent {\bf Results } The scatter plots in the top row of Figure \ref{boostingcompactresults.fig} show visualizations of the experiment above. The horizontal axes show post-processed ensemble size and the vertical axes show test error; the spread in each scatter plot is due to the 5-fold CV. We observe that the best models pruned by each of the 3 methods have similar test errors, however, \textsc{ForestPrune} produces more compact models compared to baseline and LASSO post-processing. This result is consistent across all datasets in the experiment. The bottom left plot in Figure \ref{boostingcompactresults.fig} compares the distribution of the sizes of the best models pruned. On average the best model pruned by \textsc{ForestPrune} is around 1000 nodes smaller than the best models pruned by the competing methods.

The bottom right plot in Figure \ref{boostingcompactresults.fig} shows the results of our size budget boosting experiment. The vertical axis shows the \% increase in test error between the competing methods and \textsc{ForestPrune}, and the horizontal axis shows the maximum number of nodes allowed in the pruned ensemble. We observe that for pruning sparse ensembles, with a size budget of 50 nodes, \textsc{ForestPrune} outperforms the competing algorithms. The sparse ensembles produced by the strongest competing method, BSTS, perform around \textbf{43\%} worse than the ensembles pruned by \textsc{ForestPrune}. Given such a tight node budget, the competing algorithms can only select a few trees. \textsc{ForestPrune} is more flexible and can trim and then select multiple shallow trees.

When the budget is increased to 1000 nodes, \textsc{ForestPrune} still significantly outperforms LASSO and baseline post-processing but performs similarly to BSTS. Finally, for a budget of 5000 nodes, LASSO, BSTS, and \textsc{ForestPrune} all perform comparably. This is expected since nearly the entire ensemble can be selected for this budget. We conclude that \textsc{ForestPrune} is the best performing algorithm for pruning sparse models from boosted tree ensembles, and matches our strong BSTS benchmark for selecting larger subensembles.

Code for our experiments, along with a Python implementation of \textsc{ForestPrune} can be found in our project GitHub repository.

\subsection{Interpretability Case Study}

\begin{figure}[h]
\centering
\begin{subfigure}[b]{ 0.45\textwidth}
\centering
   \includegraphics[width=\linewidth]{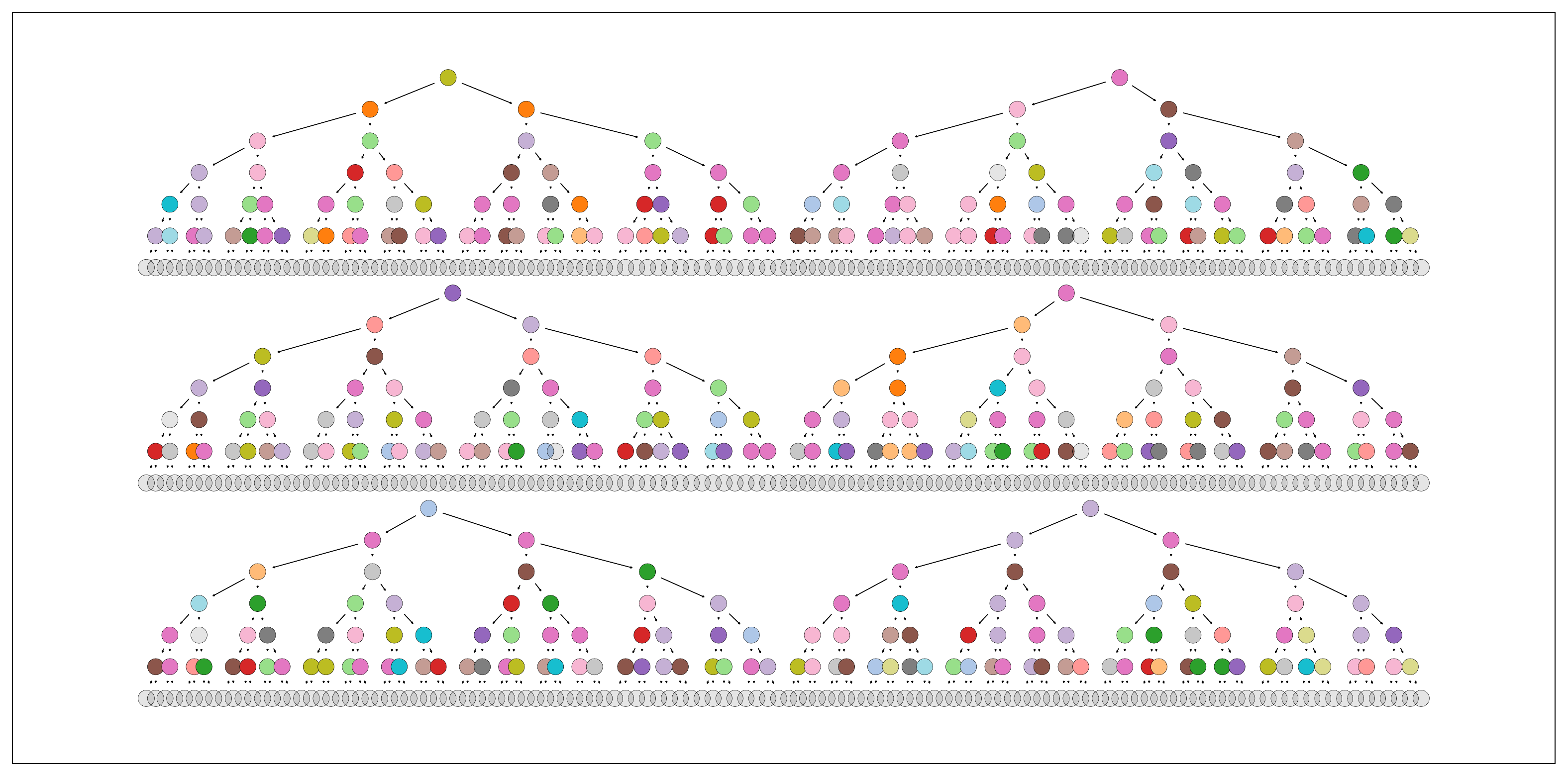}
   \caption{ A sample of full depth trees in the original ensemble.}
   \label{census1.fig}
\end{subfigure}
\centering
\begin{subfigure}[b]{0.45\textwidth}
\centering
   \includegraphics[width=\linewidth]{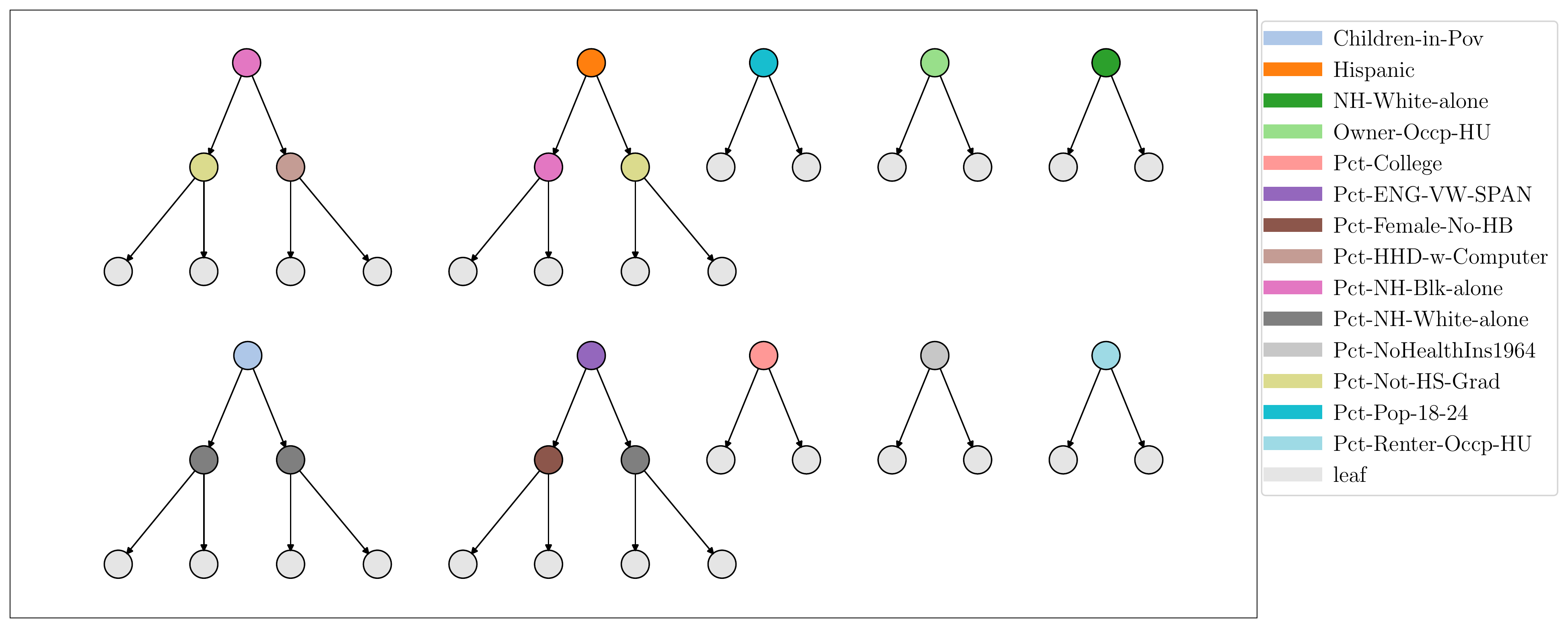}
   \caption{\textsc{ForestPrune} trims and selects 10 shallow trees.}
   \label{census2.fig} 
 \end{subfigure}
 \caption{\textsc{ForestPrune} improves the interpretability of a random forest model used to predict census response rates.}
\end{figure}

We conclude with a case study to showcase how \textsc{ForestPrune} prunes tree ensembles into interpretable models. Building off of the work by \cite{ibrahim2021predicting} we use the Census Planning Database (50,000 rows and 295 features) to predict decennial census response rates at the tract level. A modeling competition by the Census Bureau found that tree ensembles work well for this problem, but are uninterpretable and fail to produce actionable insights towards improving response rates \citep{erdman2017low}.

We start with a random forest of 500 depth 6 trees; the test RMSE of the model is \textbf{7.54\%} and all 295 features are used. A sample of 6 out of the 500 trees is presented in Figure \ref{census1.fig}. This model is difficult to interpret since the depth of each tree makes it impossible to track the relationships between the features. We use \textsc{ForestPrune} with $\alpha = 15$ to prune this random forest and use best subset polishing (\S\ref{polish.section}) to select 10 trees for further analysis. Figure \ref{census2.fig} presents this sparse model which uses just 14 out of the 295 features. The pruned ensemble achieves a test RMSE of \textbf{8.47\%}; for a small increase in test error, \textsc{ForestPrune} extracts a much more interpretable model. The model post-processed by \textsc{ForestPrune} performs much better than a single decision tree, which achieves a test RMSE of \textbf{10.6\%}.

We can examine the structure of each tree in Figure \ref{census2.fig} to understand the relationships between the features and the response. The single split, main effect trees with the features \textit{Pct-College}, \textit{Pct-Pop-18-24}, \textit{Pct-Renter-Occup-HU}, and \textit{Owner-Occup-HU}, reveal that census tracts with many non-permanent residents (renters and college students) have lower response rates. The depth 2 trees show pairwise feature interactions between ethnicity/race, education, and poverty levels and we can examine these trees to determine underserved communities have lower census response rates as well.  \textsc{ForestPrune} extracts a compact transparent model from the full tree ensemble with nominal performance loss.

\section{CONCLUSION}
In this work, we develop \textsc{ForestPrune}, an optimization framework to prune depth layers from trees in an ensemble. We propose a CBCD method with local search to compute high-quality solutions to the optimization problems formulated under this framework. Our specialized optimization algorithm is computationally efficient; the per-iteration cost of our method is linear in the number of training samples and our main algorithm converges in a few seconds for medium-sized problems. The framework only contains a single hyperparameter to tune, the regularization parameter $\alpha$, and we can leverage warm start continuation to rapidly compute the entire regularization path. Our results show that \textsc{ForestPrune} can drastically reduce the size of bagging and boosting ensembles with nominal performance loss compared to competing algorithms. Finally, \textsc{ForestPrune} can produce interpretable models by pruning ensembles shallow enough for practitioners to examine by hand, increasing the transparency of tree-based models.

\section{ACKNOWLEDGMENTS} We thank the anonymous reviewers for their comments that helped us improve the paper. This research is funded in part by a grant from the Office of Naval Research (ONR-N00014-21-1-2841).

\bibliographystyle{plainnat}
\bibliography{ref}

\end{document}